\documentclass[sigconf]{acmart}

\usepackage{multirow}
\AtBeginDocument{%
  }

\setcopyright{acmcopyright}
\copyrightyear{2023}
\acmYear{2023}
\acmDOI{XXXXXXX.XXXXXXX}

\acmConference[MM'23]{Make sure to enter the correct
  conference title from your rights confirmation emai}{October 29- November 3,
  2023}{Canada Ottawa}
\acmPrice{15.00}
\acmISBN{978-1-4503-XXXX-X/18/06}

\begin{document}

\title{
Instance-Variant Loss with Gaussian RBF Kernel for 3D Cross-modal Retrieval
}
% CSIVG: Cross-modal Softmax with Instance-Variant and Gaussian Kernel for 3D Cross-modal Retrieval}

\author{Zhitao Liu}
\authornote{Both authors contributed equally to this research.}
\email{zl425uestc@gmail.com}
\orcid{0000-0003-3499-4157}
\affiliation{%
  \institution{University of Electronic Science and Technology of China}
  \city{Chengdu}
%   \state{Sichuan}
  \country{China}
  \postcode{611731}
}

\author{Zengyu Liu}
\authornotemark[1]
\email{lzystan2021@163.com}
\affiliation{%
  \institution{University of Electronic Science and Technology of China}
  \city{Chengdu}
%   \state{Sichuan}
  \country{China}
  \postcode{611731}
}

\author{Jiwei Wei}
\email{mathematic6@gmail.com}
\affiliation{%
  \institution{University of Electronic Science and Technology of China}
  \city{Chengdu}
%   \state{Sichuan}
  \country{China}
  \postcode{611731}
}

\author{Guan Wang}
\email{wangguan12621@gmail.com}
\affiliation{%
  \institution{University of Electronic Science and Technology of China}
  \city{Chengdu}
%   \state{Sichuan}
  \country{China}
  \postcode{611731}
}

\author{Zhenjiang Du}
\email{zhenjiang@std.uestc.edu.cn}
\affiliation{%
  \institution{University of Electronic Science and Technology of China}
  \city{Chengdu}
%   \state{Sichuan}
  \country{China}
  \postcode{611731}
}

\author{Ning Xie}
\authornote{Corresponding Author}
\email{seanxiening@gmail.com}
\affiliation{%
  \institution{University of Electronic Science and Technology of China}
  \city{Chengdu}
%   \state{Sichuan}
  \country{China}
  \postcode{611731}
}

\author{Hengtao Shen}

\email{shenhengtao@hotmail.com}
\affiliation{%
  \institution{University of Electronic Science and Technology of China}
  \city{Chengdu}
%   \state{Sichuan}
  \country{China}
  \postcode{611731}
}

\renewcommand{\shortauthors}{Trovato et al.}

\begin{abstract}
3D cross-modal retrieval is gaining attention in the multimedia community. Central to this topic is learning a joint embedding space to represent data from different modalities, such as images, 3D point clouds, and polygon meshes, to extract modality-invariant and discriminative features. Hence, the performance of cross-modal retrieval methods heavily depends on the representational capacity of this embedding space. Existing methods treat all instances equally, applying the same penalty strength to instances with varying degrees of difficulty, ignoring the differences between instances. This can result in ambiguous convergence or local optima, severely compromising the separability of the feature space.
To address this limitation, we propose an Instance-Variant loss to assign different penalty strengths to different instances, improving the space separability. Specifically, we assign different penalty weights to instances positively related to their intra-class distance. Simultaneously, we reduce the cross-modal discrepancy between features by learning a shared weight vector for the same class data from different modalities. By leveraging the Gaussian RBF kernel to evaluate sample similarity, we further propose an Intra-Class loss function that minimizes the intra-class distance among same-class instances. Extensive experiments on three 3D cross-modal datasets show that our proposed method surpasses recent state-of-the-art  approaches.
\end{abstract}





\maketitle

\section{Introduction}

As 3D models become increasingly prevalent in CAD, VR/AR, and autonomous driving applications, the efficient and accurate retrieval of 3D models has gained growing attention within the multimedia community.
This area has received widespread interest as the foundation for numerous downstream tasks, such as robot navigation, scene understanding, 3D modeling, and animation~\cite{han2019image}.
3D cross-modal retrieval aims to reduce cross-modal discrepancy and learn modality-invariant (minimizing intra-class distance) and discriminative features (maximizing inter-class distance) among multi-modal data.
In comparison to 2D cross-modal retrieval (image-text retrieval~\cite{10054421,DBLP:conf/cvpr/ZhangMWZ22}, sketch-based image retrieval~\cite{DBLP:conf/cvpr/SainBYXS21,DBLP:conf/mm/TianXWSL21}), 3D cross-modal retrieval has to consider the representation and structure of 3D models and utilizes more multi-modal data, including images, point clouds, meshes, and multi-view grayscale images for 3D models as query domains for retrieval.

\begin{figure}
    \centering
    \includegraphics[width= \columnwidth]{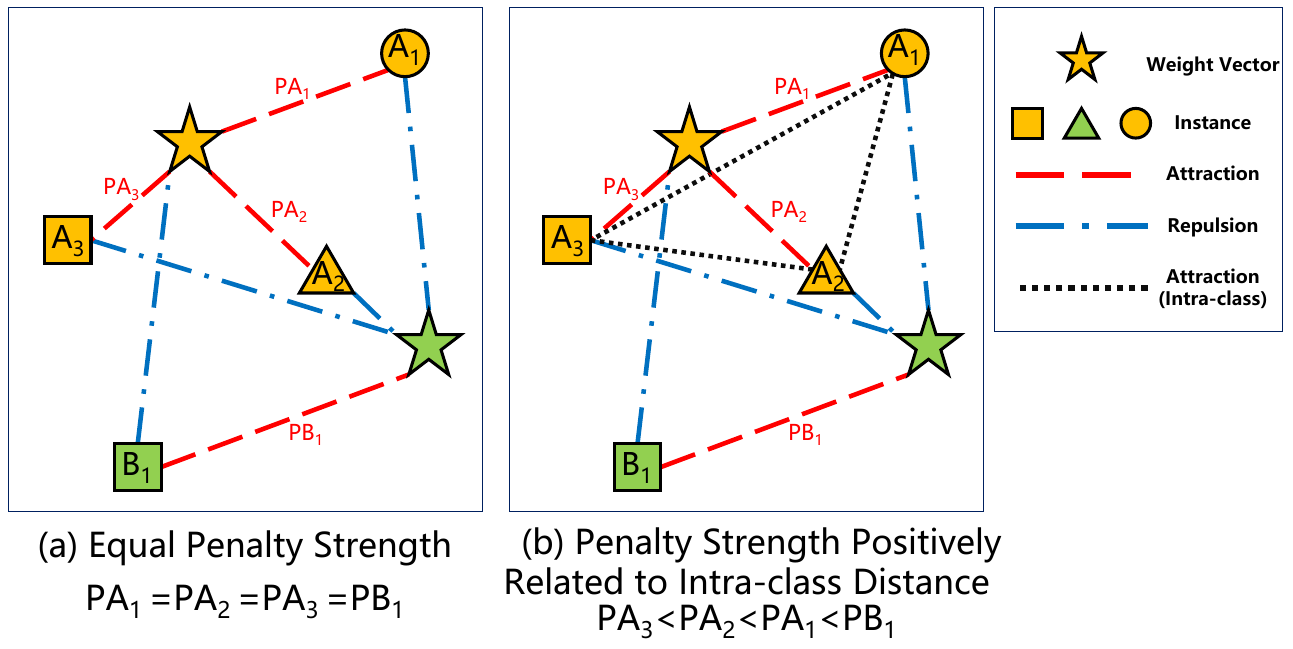}
    \caption{Illustration of Instance-Variant loss with Gaussian RBF kernel (Intra-Class loss) for 3D cross-modal retrieval. Points with the same shape are from the same modality. Different colors represent different categories. $PA_1$ and $PB_1$ denote the penalty strengths of each instance. The lengths of the red and blue dashed lines represent the intra-class and inter-class distances, respectively, acting as attraction and repulsion between instances and weight vectors. The greater the penalty strengths, the stronger the attraction and repulsion.
(a): All instances have the same penalty strength; (b): Different instances have varying penalty strengths positively related to intra-class distances. The Intra-Class loss can bring intra-class instances closer together.}
    \label{fig:fig1}
\end{figure}
% \textcolor{red}{[Add some references for these applications ?]} \textcolor{blue}{okay, I will add some, also prof suggest me to put this sentence above before 3D cross-modal} \textcolor{red}{[Yes, I think that will make sense.]}
% The key challenge of 3D cross-modal retrieval is to reduce the large cross-modal discrepancy between 3D models and images. 
% Existing cross-modal retrieval methods focus on learning a common embedding space to bridge the heterogeneous gap, enabling comparing features from different modalities. 
% The common embedding space can minimize the intra-class distance from the same category and maximize the inter-class distance from a different category.
% Recently, researchers have used contrastive loss~\cite{DBLP:conf/iccv/Lin0WLJZ021,hu2022feature}, cross-modal center loss~\cite{DBLP:conf/cvpr/JingVTT21}, softmax cross-entropy loss~\cite{DBLP:journals/tip/LiangDW21}, and other contrastive learning and metric learning methods to learn the common embedding space that characterizes multi-modal data features. 
% Jing et al.~\cite{DBLP:conf/cvpr/JingVTT21} use cross-modal center loss and cross-entropy loss to learn a common embedding space for all modality data by computing the class center (feature mean) of different classes to minimize the intra-class distance. 
% Liu et al.~\cite{hu2022feature} use the method of contrastive learning to force the inter-modal centroid alignment to reduce the modal discrepancy.

The key challenge in 3D cross-modal retrieval is reducing the substantial cross-modal discrepancy between 3D models and images.
Existing cross-modal retrieval methods focus on learning a common embedding space to bridge the heterogeneous gap, enabling comparing features from different modalities.
The common embedding space aims to minimize the intra-class distance within the same category and maximize the inter-class distance between different categories.
Recently, researchers have employed contrastive loss~\cite{DBLP:conf/iccv/Lin0WLJZ021,hu2022feature}, cross-modal center loss~\cite{DBLP:conf/cvpr/JingVTT21}, softmax cross-entropy loss~\cite{DBLP:journals/tip/LiangDW21}, and other contrastive learning and metric learning methods to learn the common embedding space that characterizes multi-modal data features.
% To minimize the intra-class distance, researchers use cross-modal center and cross-entropy loss to learn a common embedding space for all modality data by calculating the class center (mean of the feature) for different classes. They also employ contrastive learning to enforce inter-modal centroid alignment, which reduces modal discrepancy between various data modalities. This approach helps to create a more unified representation of data from different sources, ultimately enhancing the performance of cross-modal retrieval tasks.
To minimize the intra-class distance, researchers use the cross-modal center and cross-entropy loss to learn a common embedding space for all modality data by calculating the class center (mean of the feature) for different classes~\cite{DBLP:conf/cvpr/JingVTT21}. Some also employed contrastive learning to enforce inter-modal centroid alignment, reducing the modal discrepancy between various data modalities~\cite{hu2022feature}.

Although these methods have made significant progress, most still treat all instances of all classes from all modalities equally and overlook differences among instances, leading to ambiguous convergence and suboptimal performance~\cite{DBLP:journals/pami/WeiYXZS22}. 
Cross-modal retrieval tasks improve the network's representation ability in the common space by optimizing the margin between the intra-class and inter-class distances. 
% For instance, consider the cosine distance commonly used in the retrieval field. Take two different modality instances, $A_1$ and $A_2$, of the same class; $A_1$ instance $\{$\textit{intra-class distance=0.4}, \textit{inter-class distance=0.6}$\}$ and $A_2$ instance $\{$\textit{intra-class distance=0.1}, \textit{inter-class distance=0.3}$\}$. The margin for both instances is 0.2, contributing the same to the network (from the perspective of loss optimization); however, the inter-class distance of $A_2$ is only 0.3, which is even smaller than the intra-class distance of $A_1$. 
% The contribution of different instances to the network varies considerably, as illustrated in figure \ref{fig:fig1}.
% % If the network applies the same penalty strength to instances, $A_1$ and $A_2$, ambiguous convergence (local optima) will severely compromise the feature space's separability, as illustrated in figure \ref{fig:fig1}. 
% Instance A2 can be considered a hard or ambiguous sample that resides close to the decision boundary or exhibits characteristics of both classes. As a result, the model might have difficulty learning an optimal decision boundary or feature representation that can effectively separate instances from classes A and B. 
As illustrated in Fig. \ref{fig:fig1}, the contribution of different instances to the network varies considerably. Applying equal penalty strength to instances with varying degrees of difficulty can result in ambiguous convergence or local optima, which can severely compromise the separability of the feature space.
Furthermore, the inherent data representation and structural differences between 3D models and images make learning a separable embedding space challenging. This significant modality discrepancy will make it hard for the network to learn modality-invariant features from the data.
As a result, the retrieval performance in 3D cross-modal tasks often suffers from ambiguous convergence. Therefore, it is crucial to propose an effective instance-level penalized-strength weighted loss for 3D cross-modal retrieval tasks. This approach would consider the varying contributions of instances to the network and assign different penalty strengths to different instances.

To address the aforementioned issues, we propose the Instance-Variant loss for 3D cross-modal retrieval. The most intuitive motivation is to assign different penalty strengths to different instances. By adding a learnable weight coefficient, positively related to the instance's intra-class distance, to the softmax output. We ensure that instances more challenging to distinguish (with larger intra-class distances) receive more severe penalty strengths, thereby improving the network's effectiveness. Simultaneously, because the penalty strength is related to the intra-class distance of instances, the network will also adapt to the problems of data distribution imbalance and multimodal optimization imbalance. Unlike the classical softmax loss, we add different penalty strengths for different instances and learn a common weight vector for all modalities. This approach will effectively alleviate the modal optimization imbalance problem and reduce the discrepancy between modalities.

Under the constraint of softmax loss, we can map all modality data onto a shared unit hypersphere~\cite{DBLP:conf/cvpr/LiuWYLRS17} by normalizing the weight vector and instance features. The unit hypersphere can better minimize intra-class distance and maximize inter-class distance. It also helps reduce the discrepancy between various data sources, enhance the semantic Information of the learned data, and improve the robustness of retrieval tasks. Inspired by the Gaussian RBF kernel function~\cite{borodachov2019discrete}, we propose an Intra-Class loss, which aims to minimize the distance between all samples of the same class on the unit hypersphere. Unlike Centerloss, our Intra-Class loss does not require finding class centers. The essence of the Gaussian kernel is to measure the similarity between samples. Similar samples can be better clustered together in a space that describes similarity and then become linearly separable. Using Intra-Class loss can minimize the intra-class distance and evaluate the multi-modal shared unit hypersphere (embedding space) learned by the network.

By utilizing the proposed Instance-Variant loss and Intra-Class loss, the network can assign distinct penalty strengths to different instances while minimizing the intra-class distance, effectively reducing the inseparability of the feature space caused by ambiguous convergence. 
% \textit{The proposed CSIV and intra-class loss can be jointly trained with other loss functions to optimize the feature extraction or mapping network.}
To validate the effectiveness of the proposed Instance-Variant and Intra-Class loss, we jointly train the framework with the cross-entropy loss function for the 3D cross-modal retrieval task, aiming to extract modality-invariant and discriminative features. Our approach significantly outperforms recent state-of-the-art methods in 3D cross-modal and uni-modal retrieval tasks. The primary contributions of this paper can be summarized as follows:

\begin{itemize}
\item We propose the Instance-Variant loss, which effectively assigns different penalty strengths to different instances, enhancing the network's effectiveness by focusing on more challenging instances and promoting better feature space separability.
\item We introduce the Intra-Class loss based on the Gaussian RBF kernel function, aiming to minimize the intra-class distance among all instances in the shared embedding space. This approach minimizes the intra-class distance and evaluates the multi-modal shared embedding space learned by the network.
\item The proposed Instance-Variant loss learns a shared weight vector for data from different modalities, effectively mitigating cross-modal discrepancy and enhancing cross-modal retrieval performance. 
% \item The proposed CSIV loss and Intra-class loss can be jointly trained with other loss functions, optimizing the feature extraction or mapping network and enabling the extraction of modality-invariant and discriminative features.
\item Our approach significantly outperforms the recent state-of-the-art methods on three datasets (Pix3D, ModelNet40, MI3DOR) for 3D cross-modal and uni-modal retrieval tasks. This demonstrates the effectiveness of the proposed Instance-Variant loss and Intra-Class loss.
\end{itemize}

\section{Related Works}
% In this section, we briefly review two related topics: 3D Cross-modal Retrieval and Metric Learning.

\textbf{3D Cross-modal Retrieval.} 
There has been growing interest in 3D Cross-modal retrieval algorithms, including image-based 3D shape retrieval (IBSR), 3D model-based shape retrieval, and 3D Cross-modal mutual retrieval.
Image-based approaches represent a 3D shape as a set of 2D views captured from pre-defined viewpoints of the 3D shape~\cite{DBLP:journals/tcsv/HuangYLLL22,DBLP:journals/tcsv/ZhouLZCLL22,DBLP:journals/tmm/XuHWZG20}. The advantage of using a set of 2D view representations is that they can directly employ the existing powerful CNNs for feature extraction~\cite{DBLP:journals/pami/Liu0JJLL023,DBLP:conf/iccv/SuMKL15} and reduce the domain gap between 3D models and images. 
% Moreover, the domain gap between 2D views and images is smaller than point clouds, meshes, and other conventional 3D model representation methods.
Lin et al.\cite{DBLP:conf/iccv/Lin0WLJZ021} used contrastive learning to realize instance-level 3D shape retrieval based on a single image. So far, great progress has been made in IBSR tasks. However, there are still some limitations, such as the queried domain of this task being limited to images and the lack of global spatial and local geometric information on the 3D model.
Model-based shape retrieval typically represents 3D models in the form of polygon meshes~\cite{DBLP:journals/tog/HanockaHFGFC19,DBLP:conf/aaai/FengFYZG19} and point clouds~\cite{DBLP:conf/cvpr/QiSMG17,DBLP:journals/tog/WangSLSBS19,DBLP:conf/nips/QiYSG17}. 
% Some representative methods include:
PointNet~\cite{DBLP:conf/cvpr/QiSMG17} developed a point-wise operation and a symmetric function to solve the permutation variance issue. 
% It also introduced T-Net to address rotation and translation issues.
DGCNN~\cite{DBLP:journals/tog/WangSLSBS19} proposed a dynamic graph convolution neural network with EdgeConv using K nearest neighbor points.
MeshNet~\cite{DBLP:conf/aaai/FengFYZG19} and MeshCNN~\cite{DBLP:journals/tog/HanockaHFGFC19} were designed to learn features directly from the mesh by modeling the geometric relations of the mesh faces of the object.
The advantage of model-based approaches is they can explore the global spatial and local geometric information of 3D models to obtain representative 3D descriptors. However, they have primarily focused on uni-modal retrieval without further exploring cross-modal retrieval. 

3D Cross-modal mutual retrieval comprehensively considers the advantages and disadvantages of the aforementioned methods. They leverage feature extraction networks like DGCNN, ImageNet, and MeshNet to extract features from point clouds, images, and polygon meshes. By measuring the intra-class and inter-class distances in a shared embedding space, cross-modal mutual retrieval can achieve mutual retrieval from any two modals.
Jing et al.~\cite{DBLP:conf/cvpr/JingVTT21} pioneered the field of 3D cross-modal (mutual) retrieval, using cross-modal center loss to reduce the discrepancy between modalities and minimize the intra-class distance of samples. Chen et al.\cite{DBLP:conf/bmvc/ChenJLT021} utilized multimodal contrastive prototype loss to accomplish semi-supervised 3D cross-modal retrieval.
3D cross-modal mutual retrieval has far-reaching implications. It enables cross-modal retrieval from real (projected) images, point clouds, and mesh data. Using 3D cross-modal (mutual) retrieval technology, researchers can quickly find corresponding texture materials for 3D models and rapidly convert between different 3D model representation methods. 
This paper aims to use the proposed Instance-Variant loss and Intra-Class loss to achieve cross-modal mutual retrieval between real images and 3D models.

\textbf{Metric learning.} 
Metric learning focuses on learning a distance metric function that encourages semantic relevant instances close to each other. In previous literature, numerous metric learning approaches~\cite{DBLP:conf/cvpr/WangHHDS19,DBLP:conf/cvpr/ZhengL020,DBLP:conf/cvpr/SunCZZZWW20} have been developed for various tasks. 
Wang et al.\cite{DBLP:conf/cvpr/WangHHDS19} proposed a multi-similarity loss for collecting and weighting informative pairs. 
% Zheng et al.\cite{DBLP:conf/cvpr/ZhengL020} developed a metric learning method named MDL-ALA, which aims to maximize the generalization ability of the learned metric. 
Sun et al.\cite{DBLP:conf/cvpr/SunCZZZWW20}introduced a circle loss to weight different similarity scores.
However, the above approaches are developed for unimodal retrieval tasks, which usually cannot accurately capture the relationship of cross-modal components with the modality gap. 
% Recently, few metric learning approaches have been implemented for cross-modal retrieval.
Frome et al.\cite{DBLP:conf/nips/FromeCSBDRM13} attempted to project images and sentences into a common embedding space, and an unweighted triplet loss was used to encourage relevant semantic instances to cluster together. 
Wei et al.\cite{DBLP:journals/pami/WeiYXZS22,DBLP:conf/mm/WeiXWW21} introduced the universal weighting metric learning framework to sample informative pairs and assign proper weight values to them based on their similarity scores.
However, the above methods can't adapt to category-level cross-modal retrieval.  Constrained by the limitations of existing datasets, it is difficult for researchers to construct sample pairs or triplets. Hence, in this paper, we develop a novel metric learning method for 3D cross-modal retrieval, which can modify its penalty strength directly according to the intra-class distance of the instance, changing its contribution to the loss function.

\section{Proposed Approach}

In this section, we will explain our proposed Instance-Variant loss and the Intra-Class loss based on the Gaussian RBF kernel function. 
In subsection 3.1, we will present the problem description and preliminaries. Subsections 3.2 and 3.3 will cover the mathematical formulation and analysis of the Instance-Variant and Intra-Class losses, respectively.
% We will present the problem description and preliminaries in subsection 3.1, followed by the mathematical 
% formulation and analysis of the Instance-Variant and Intra-Class losses in subsections 3.2 and 3.3, respectively.
% We will present the problem description and preliminaries in subsection 3.1, followed by the mathematical 
% formulation and analysis of the Instance-Variant and Intra-Class losses in subsections 3.2 and 3.3
% % .
% , respectively.

\subsection{Problem Statement and Preliminaries}
Assuming dataset $S$ contains $n*N*M$ instances, with $n$ instances of $N$ categories from $M$ modalities, the $i$-th instance $t_i$ is a set consisting of the set of modalities $s_i$ with a semantic label $y_i$ and weight vector $W_{y_i}$. Formally:

\begin{equation}
\begin{split}
&S = \{t_i\}_{i=1}^{n*N},\qquad t_i=(s_i,y_i,W{y_i}),\\ & s_i = \{x_{i}^{m}\}_{m=1}^{M},\qquad W{y_i}=\{w_{y_i}^{m}\}_{m=1}^{M}.
\end{split}
\end{equation}
% Generally, the modality instances $\{{x_{i}^{1},x_{i}^{2},\cdots,x_{i}^{m}}\}$ are in $M$ different representation spaces, and their similarities (distances) cannot be directly measured. The 3D cross-modal retrieval task aims to learn $M$ projection functions $f_m$ for each modality $m \in [1,M]$, where $v_{i}^{m}=f_m(x_{i}^{m},\theta_{m})$, $V_{y_i}^m=\{v_{1}^{m},v_{2}^{m},\cdots,v_{n}^{m}\}$.
% As a result, $V_{i}^{m}$ is a projected feature in the common representation space. To achieve good retrieval performance, we want the distance of the same-class instance to be smaller than the distance between the instance from different classes by a large margin. Thus the intra-class distance should be smaller than the inter-class.
% \begin{equation}
% \begin{split}
%     & D(V_{y_i}^{m}, V_{y_i}^{m'}) + \lambda_0 < D(V_{y_i}^{m}, V_{y_j}^{m'}) , \\
%     & D(V_{y_i}^{m}, W_{y_i}) + \lambda_0  < D(V_{y_i}^{m}, W_{y_j}) ,
% \end{split}
% \end{equation}
% here, $\lambda_0$ is a margin.

Generally, the modality instances $\{x_{i}^{1},x_{i}^{2},\cdots,x_{i}^{M}\}$ are in $M$ different representation spaces, and their similarities (distances) cannot be directly measured. The 3D cross-modal retrieval task aims to learn $M$ projection functions $f_m$ for each modality $m \in [1, M]$, where $v_{i}^{m}=f_m(x_{i}^{m},\theta_{m})$, $V_{y_i}^m=\{v_{1}^{m},v_{2}^{m},\cdots,v_{n}^{m}\}$, $\theta_m$ is a learnable parameter.
$v_{i}^{m}$ is a projected feature in the common representation space, and $V_{y_i}^{m}$ is the set of all features of the same class. 
To obtain optimal retrieval performance, we aim to ensure that the distance between same-class instances and the intra-class distance $D(V_{y_i}^{m}, W_{y_i})$ is smaller than the distance between instances from different classes and the inter-class distance $D(V_{y_i}^{m}, W_{j})$ by a significant margin $\lambda_0$.
\begin{equation}
\begin{split}
& D(V_{y_i}^{m}, V_{y_i}^{m'}) + \lambda_0 < D(V_{y_i}^{m}, V_{y_j}^{m'}) , \\
& D(V_{y_i}^{m}, W_{y_i}) + \lambda_0 < D(V_{y_i}^{m}, W_{j}) .
\end{split}
\end{equation}

To better understand the proposed Instance-Variant loss, we will first briefly review the original softmax, A-softmax~\cite{DBLP:conf/cvpr/LiuWYLRS17}, and AM-softmax~\cite{DBLP:journals/spl/WangCLL18}. The formulation of the original softmax loss is given by
\begin{equation}
\begin{aligned}
\mathcal{L}_S & =-\frac{1}{n}\frac{1}{N} \sum_{i=1}^{n*N} \log \frac{e^{W_{y_i} \cdot \boldsymbol{f}_i}}{\sum_{j=1}^N e^{W_j \cdot \boldsymbol{f}_i}} \\
& =- \frac{1}{n} \frac{1}{N} \sum_{i=1}^{n*N} \log \frac{e^{\left\|W_{y_i}\right\|\left\|\boldsymbol{f}_i\right\| \cos \left(\theta_{y_i}\right)}}{\sum_{j=1}^N e^{\left\|W_j\right\|\left\|\boldsymbol{f}_i\right\| \cos \left(\theta_j\right)}}, 
\end{aligned}
\end{equation}
where $f$ is the input of the last fully connected layer ($f_i$ denotes the $i$-th sample), and $W_j$ is the $j$-th column of the last fully connected layer. The $W{y_i} \cdot \boldsymbol{f}_i$ is also called the target logit~\cite{DBLP:conf/iclr/PereyraTCKH17} of the $i$-th sample. $W_j$ is also called the weight vector of the $j$-th category. The relationship between the weight vector and the features' mean vector (class center) is described in Figure 6 of~\cite{DBLP:conf/mm/WangXCY17}. In the A-softmax loss, the authors proposed to normalize the weight vectors (making $\left\|W_i\right\|$ to be 1) and generalize the target logit from $\left\|f_i\right\| \cos (\theta{y_i})$ to $\left\|f_i\right\| \psi (\theta_{y_i})$, where $\psi(\theta)=\frac{(-1)^k \cos (m \theta)-2 k+\lambda \cos (\theta)}{1+\lambda}$. In the AM-softmax, the authors proposed to normalize both the weight vectors and instances' features ($\left\|W_i\right\|=\left\|f_i\right\|=1$), while the $\psi (\theta_{y_i})=\cos (\theta_{y_i})-\lambda_0$.

\subsection{Instance-Variant Loss}
The softmax loss function has been extensively applied in uni-modal retrieval tasks, as it not only identifies the optimal plane to separate distinct data classes but also learns the ideal weight vector for each category. In comparison to the class center (feature mean), the weight vector is better suited for metric learning problems involving hard samples.
Nevertheless, in contrast to uni-modal retrieval tasks, cross-modal retrieval tasks must also diminish cross-modal discrepancy. We innovatively learn a shared weight vector for data across various modalities ($W{y_i}=w_{y_i}^{1}=w_{y_i}^{M}$). Throughout the weight vector's iterative process, it will acquire the features of all modal data, effectively addressing the imbalance in modal optimization. This paper assumes that the norm of $W_i$ and $f$ are normalized to 1 if not specified. Given the extracted features $\{V_{i}^m\}\quad(i\in[1,n*N],m\in[1,M])$, $f_{i}^m=V_{i}^m$. The new cross-modal softmax loss takes the following form. 

\begin{equation}
    \begin{aligned}
\mathcal{L}_{NS} & =-\frac{1}{n} \frac{1}{N} \frac{1}{M} \sum_{i=1}^{n*N} \sum_{m=1}^M \log \frac{e^{w_{y_i} \cdot \boldsymbol{f}_{i}^m}}{\sum_{j=1}^c e^{w_j \cdot \boldsymbol{f}_{i}^m}} \\
& = -\frac{1}{n} \frac{1}{N} \frac{1}{M} \sum_{i=1}^{n*N} \sum_{m=1}^M \log \frac{e^{w_{y_i} \cdot \boldsymbol{V}_{i}^m}}{\sum_{j=1}^c e^{w_j \cdot \boldsymbol{V}_{i}^m}}.
\end{aligned}
\end{equation}
To simplify subsequent derivations, we can rewrite Equation 4 as Equation 5.
\begin{equation}
\begin{aligned}
    \mathcal{L}_{NS'} & = -\frac{1}{n} \frac{1}{N} \frac{1}{M} \sum_{i=1}^{n*N} \sum_{m=1}^M \log \frac{e^\phi(\theta_{y_i}^m)}{e^{\phi(\theta_{y_i}^m)} + \sum_{j=1,j\neq {y_i}}^N e^{\eta(\theta_{j}^m)}} \\
& = \frac{1}{n} \frac{1}{N} \frac{1}{M}  \sum_{i=1}^{n*N} \sum_{m=1}^M \log(1+\sum_{j=1,j\neq {y_i}}^N e^{\eta(\theta_{j}^m) - \phi(\theta_{y_i}^m)}).
\end{aligned}
\end{equation}
To ensure the intra-class distance is smaller than the inter-class distance,
$1-\cos (\theta_{j}^m) - (1-\cos (\theta_{y_i}^m)) \geq \lambda_0,\quad \lambda_0>0$, where $\cos (\theta_{y_i}^m)-\lambda_0 \geq \cos (\theta_{j}^m)$, where $\cos (\theta_{y_i}^m)>\cos (\theta_{j}^m)$.

Thus, we can let $\eta(\theta_{j}^m) = \left\|W_j\right\|\left\|\boldsymbol{f}_{i}^m\right\| \cos (\theta_{j}^m) = \cos (\theta_{j}^m)$,

and $\phi(\theta_{y_i}^m) = \left\|W_i\right\|\left\|\boldsymbol{f}_{i}^m\right\| \cos (\theta_{y_i}^m) - \lambda_0 = \cos (\theta_{y_i}^m) - \lambda_0 $ .

If the network applies the same penalty strength to instances, ambiguous convergence (local optima) severely compromises the feature space's separability.  To address the limitation, we consider enhancing the optimization flexibility by allowing each instance to learn at its own pace, depending on its current optimization status. The Instance-Variant loss takes the following form.

\begin{equation}
    \begin{aligned}
        \mathcal{L}_{IV} &= \frac{1}{n} \frac{1}{N} \frac{1}{M}  \sum_{i=1}^{n*N} \sum_{m=1}^M 
        \frac{\sum_{j=1,j\neq {y_i}}^N e^{\eta(\theta_{j}^m) - \phi(\theta_{y_i}^m)}}
        {1+\sum_{j=1,j\neq {y_i}}^N e^{\eta(\theta_{j}^m) - \phi(\theta_{y_i}^m)}} \\
        & \times\log(1+\sum_{j=1,j\neq {y_i}}^N e^{\eta(\theta_{j}^m) - \phi(\theta_{y_i}^m)}).
    \end{aligned}
\end{equation}
Let $\Gamma=\sum_{j=1,j\neq {y_i}}^N e^{\eta(\theta_{j}^m) - \phi(\theta_{y_i}^m)}$, 
\begin{equation}
\begin{aligned}
    & \mathcal{L}_{IV} = \frac{1}{n} \frac{1}{N} \frac{1}{M}  \sum_{i=1}^{n*N} \sum_{m=1}^M
    \frac{\Gamma}{1+\Gamma} \times log (1+\Gamma) \\
    & = \frac{1}{n} \frac{1}{N} \frac{1}{M}  \sum_{i=1}^{n*N} \sum_{m=1}^M log(1+\Gamma)^{\frac{\Gamma}{1+\Gamma}} \\
    & \mathcal{L}_{NS'} = \frac{1}{n} \frac{1}{N} \frac{1}{M}  \sum_{i=1}^{n*N} \sum_{m=1}^M log(1+\Gamma).
\end{aligned}
\end{equation}

Under the same condition that factors affecting loss convergence, such as input data and optimizer, $\mathcal{L}_{IV} < \mathcal{L}_{NS'}$.
Take the partial derivative of $\Gamma$ with respect to $\theta_{y_i}^m$ ($\theta_{y_i}^m\in[0,\pi]$) and derivative of $\mathcal{L}_{IV}$ and $\mathcal{L}_{NS'}$ with respect to $\Gamma$, $\frac{\partial \Gamma} {\partial \theta_{y_i}^m} < 0$, $\frac{d\mathcal{L}_{IV}}{d\Gamma}>0,\frac{d\mathcal{L}_{NS'}}{d\Gamma}>0$. Therefore, we can get that both $\mathcal{L}_{IV}$ and $\mathcal{L}_{NS'}$ are monotonically decreasing concerning $\theta_{y_i}^m$. If we assume the $\mathcal{L}_{IV}$ and $\mathcal{L}_{NS'}$ are optimized to the same value and all training features can be perfectly classified (consistent with the L-softmax assumption~\cite{DBLP:conf/icml/LiuWYY16}).

\begin{equation}
    \begin{aligned}
        & \because \mathcal{L}_{IV} < \mathcal{L}_{NS'} , \quad \frac{d\mathcal{L}_{IV}}{d\Gamma}>0, \quad
        \frac{d\mathcal{L}_{NS'}}{d\Gamma} >0, \\& \qquad min(\mathcal{L}_{IV})=min(\mathcal{L}_{NS'}). \\
        & \therefore \Gamma_{IV} > \Gamma_{NS'}.  \qquad
         \because \frac{\partial \Gamma} {\partial \theta_{y_i}^m} < 0. \\
        & \therefore \{\theta_{y_i}^{m}\}_{IV} < \{\theta_{y_i}^{m}\}_{NS'}.
    \end{aligned}
\end{equation}

% When the $\mathcal{L}_{IV}$ and $\mathcal{L}_{NS'}$ are optimized to the same value and all training features can be perfectly classified, the $\theta_{y_i}^{m}$ of $\mathcal{L}_{IV}$ will be smaller than the one of $\mathcal{L}_{NS'}$, which means $\mathcal{L}_{IV}$ will have better performance than $\mathcal{L}_{NS'}$.  
% During the training optimization process, it will receive a stronger penalty when an instance is difficult to distinguish (i.e., the intra-class distance is large or the inter-class distance is small). Compared to the traditional softmax loss, we introduce the concept of hard negative mining. Moreover, we assess both the intra-class and inter-class distances to define hard instances, effectively enhancing the separability of the feature space. We can also use the hyperparameter $\tau$ to scale $\frac{\Gamma}{1+\Gamma}$, adapting to different data distributions in various datasets. Simultaneously, researchers can improve the retrieval performance of the model by modifying $\phi(\theta_{y_i}^m)$ and $\eta(\theta_{j}^m)$. The proposed Instance-Variant loss can take the following form.

When $\mathcal{L}{IV}$ and $\mathcal{L}{NS'}$ are optimized to the same value and all training features can be perfectly classified, the $\theta_{y_i}^{m}$ of $\mathcal{L}{IV}$ will be smaller than the one of $\mathcal{L}{NS'}$, which means $\mathcal{L}{IV}$ will have better performance than $\mathcal{L}{NS'}$ (Proved in Appendix).
During the training optimization process, an instance will receive a stronger penalty when it is difficult to distinguish (the intra-class distance is large or the inter-class distance is small). Compared to the traditional softmax loss, we introduce the concept of hard negative mining. Moreover, we assess both the intra-class and inter-class distances to define hard instances, effectively enhancing the separability of the feature space. We can also use the hyperparameter $\tau$ to scale $\frac{\Gamma}{1+\Gamma}$, adapting to different data distributions in various datasets. Simultaneously, researchers can improve the retrieval performance of the network by only modifying $\phi(\theta_{y_i}^m)$ and $\eta(\theta_{j}^m)$. The proposed Instance-Variant loss can take the following form.
\begin{equation}
    \begin{aligned}
        \mathcal{L}_{IV} &= \frac{1}{n} \frac{1}{N} \frac{1}{M}  \sum_{i=1}^{n*N} \sum_{m=1}^M
    {(\frac{\Gamma}{1+\Gamma})}^\tau \times log (1+\Gamma). \\
    & \Gamma=\sum_{j=1,j\neq {y_i}}^N e^{\eta(\theta_{j}^m) - \phi(\theta_{y_i}^m)} \\
    & \eta(\theta_{j}^m) =  \cos (\theta_{j}^m)/\omega, \quad \phi(\theta_{y_i}^m) = (\cos (\theta_{y_i}^m) - \lambda_0)/\omega,
    \end{aligned}
\end{equation}
where $\lambda_0$ and $\tau$ are hyperparameter, $\omega$ is the temperature coefficient~\cite{hinton2015distilling} for softmax.
\subsection{Gaussian RBF Intra-Class loss}
\begin{figure*}
    \centering
    \includegraphics[width=\textwidth]{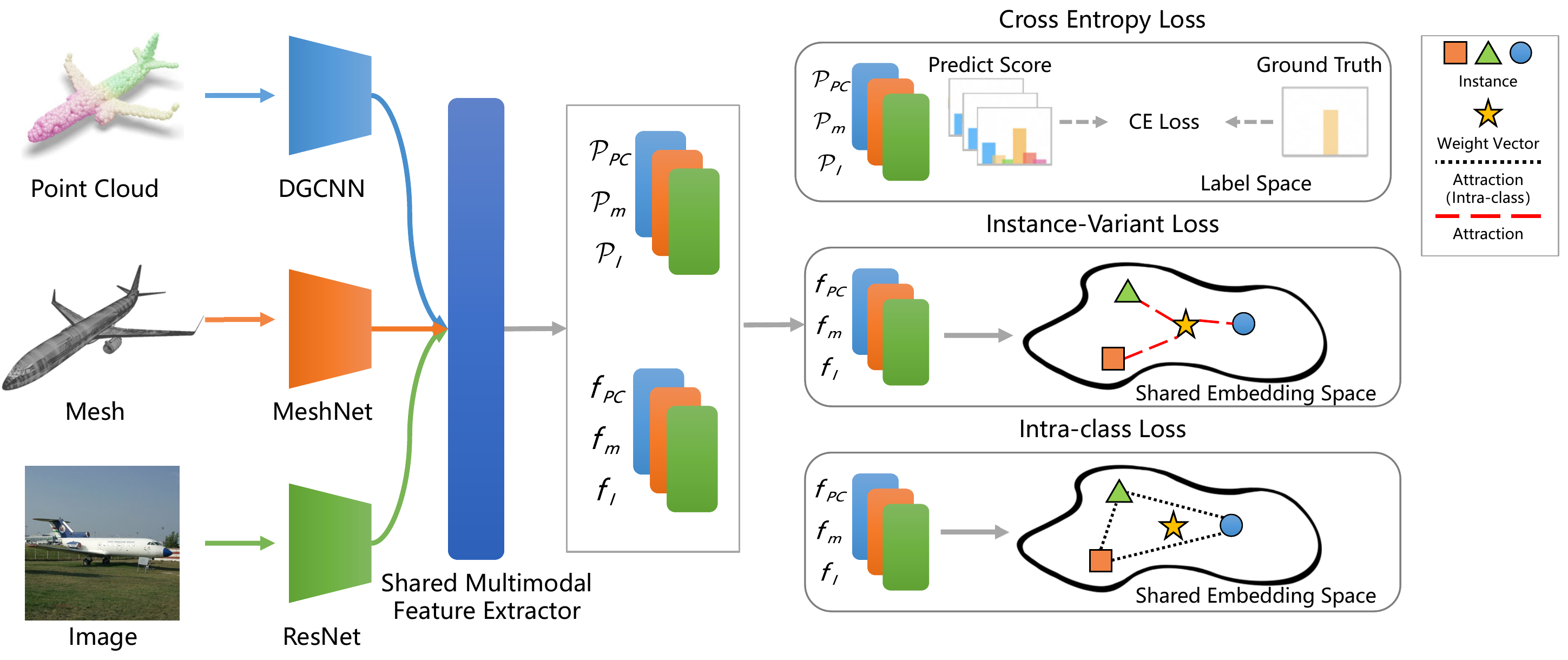}
    \caption{
    % An overview of the proposed framework for 3D cross-modal retrieval task. Mesh, point cloud, and image features are extracted by MeshNet, DGCNN, and ResNet, respectively, then projected to a common space via a shared multimodal feature extractor. The cross-entropy in the label space is used to learn discriminative features, and Instance_Variant loss in the shared embedding space (unit hypersphere) is used to learn discriminative and modality-invariant features, while the Intra-class will learn the modality-invariant features.
    An overview of our proposed framework for the 3D cross-modal retrieval task. Cross-entropy loss is employed in the label space. In the shared embedding space, Instance-Variant loss is utilized to learn discriminative and modality-invariant features. Concurrently, the Intra-class loss focuses on learning modality-invariant features, further enhancing the performance of our proposed framework. Different colors and shapes represent instances from different modalities of the same class.
    }
    \label{fig:fig2}
\end{figure*}
Under the constraint of the softmax loss, we can map all modality data onto a shared unit hypersphere by normalizing the weight vector and instance features.
We aim for the intra-class metric to be asymptotically correct and empirically reasonable with a finite number of points. Also, the essence of the Gaussian RBF kernel is to measure the similarity between samples.
Similar samples can be better clustered together in a space that describes their similarity, and subsequently become linearly separable.
We consider the Gaussian Radial Basis Function (RBF) kernel to achieve this.
\begin{equation}
\mathcal{K}_t(x,x')\triangleq e^{-t\parallel x-x'\parallel _{2}^{2}}=e^{2t\cdot x^{\top}x'-2t}, \quad t>0,
\end{equation}
and define the Intra-Class loss as the negative log-likelihood of the average pairwise Gaussian RBF:

\begin{equation}
\begin{aligned}
        \mathcal{L} _{IC} & = - \frac{1}{N}\sum_{y_{i}=1}^N{\frac{1}{n*M} \log(\sum_{i=1,j=1,i\ne j}^{n*M}{\mathcal{K}_t(x,x'))}} \\
        & =- \frac{1}{N}\sum_{y_{i}=1}^N{\frac{1}{n*M} \log(\sum_{i=1,j=1,i\ne j}^{n*M}{e^{-t\left\| x_{i}^{y_i}-x_{j}^{y_i} \right\| _{2}^{2}}})}.
\end{aligned}
\end{equation}
The objective of this loss function is to minimize the distance between any two instances of the same class from different modalities, thereby reducing the intra-class distance among all data within the same class. Compared to existing Intra-Class loss functions~\cite{DBLP:journals/tcsv/XiaoLLDL22,DBLP:conf/cvpr/JingVTT21}, our Intra-Class loss straightforwardly minimizes the distance between two instances without learning class centers, reducing the network's complexity and the cross-modal discrepancy. We will discuss more mathematic characters in the appendix.
 % The Intra-Class loss function aims to minimize the distance between instances of the same class from different modalities, reducing intra-class distance and cross-modal discrepancy. Unlike existing Intra-Class loss functions, ours straightforwardly minimizes the distance between instances without learning class centers, simplifying the network model. We'll discuss more mathematical characteristics in the appendix.
 
% By employing negative log-likelihood, we alter the property of average pairwise Gaussian RBF being tightly related to the uniform distribution. As the distance between instances of the same class decreases, the loss will also decrease. Thus, the Intra-Class loss is closely related to the distance between instances. At the same time, the original properties of the Gaussian RBF kernel still hold, such as for $M(S_d)$, the set of Borel probability measures on $S_d$, $\sigma_d$ is the unique solution~\cite{DBLP:conf/icml/0001I20} of: $ \min_{\mu \in \mathcal{M}\left(\mathcal{S}^d\right)} \int_x \int_{x'} \mathcal{K}_t(x, x') \mathrm{d} x \mathrm{d} x'.$ This property ensures that our loss function while minimizing the distance between instances of a certain class, will not be affected by or influence the data of other classes.
% \begin{equation}
% \min _{\mu \in \mathcal{M}\left(\mathcal{S}^d\right)} \int_x \int_{x'} \mathcal{K}_t(x, x') \mathrm{d} x \mathrm{d} x'.
% \end{equation}
% This property ensures that our loss function while minimizing the distance between instances of a certain class, will not be affected by or influence the data of other classes.

\begin{table*}[]
\caption{Performance of 3D uni-modal and cross-modal retrieval task on ModelNet40, MI3DOR, Pix3D-9 (original Pix3D) and Pix3D-4 datasets regarding mAP. Comparison with the state-of-the-art method CMCL~\cite{DBLP:conf/cvpr/JingVTT21}. When the target or source is from the image domain, the results are reported for multi-view images (only for ModelNet40): 1,2,4 views denoted by v1, v2, and v4.}
\centering
\begin{tabular}{cc|ccccccc|cc|cc|cc}
\hline
\multicolumn{2}{c|}{\textbf{Datasets}}                                 & \multicolumn{7}{c|}{\textbf{ModelNet40}}                                                 & \multicolumn{2}{c|}{\textbf{MI3DOR}} & \multicolumn{2}{c|}{\textbf{Pix3D-9}} & \multicolumn{2}{c}{\textbf{Pix3D-4}} \\ \hline
\multicolumn{1}{c|}{\multirow{2}{*}{Source}} & \multirow{2}{*}{Target} & \multicolumn{2}{c}{mAP-v1} & \multicolumn{2}{c}{mAP-v2} & \multicolumn{3}{c|}{mAP-v4}    & \multicolumn{2}{c|}{mAP}             & \multicolumn{2}{c|}{mAP}              & \multicolumn{2}{c}{mAP}              \\ \cline{3-15} 
\multicolumn{1}{c|}{}                        &                         & CMCL    & Ours             & CMCL    & Ours             & CMCL  & DSCMR~\cite{DBLP:conf/cvpr/ZhenHWP19} & Ours           & CMCL         & Ours                  & CMCL          & Ours                  & CMCL         & Ours                  \\ \hline
\multicolumn{1}{c|}{Image}                   & Image                   & 82.06   & \textbf{83.55}   & 86.00   & \textbf{88.13}   & 90.23 & 82.31 & \textbf{90.50} & 75.26        & \textbf{77.24}        & 72.97         & \textbf{77.99}        & 85.97        & \textbf{91.38}        \\
\multicolumn{1}{c|}{Image}                   & Mesh                    & 85.58   & \textbf{86.20}   & 87.31   & \textbf{88.64}   & 89.59 & 77.30 & \textbf{89.94} & 78.24        & \textbf{79.78}        & 73.79         & \textbf{81.64}        & 86.78        & \textbf{90.86}        \\
\multicolumn{1}{c|}{Image}                   & Point                   & 85.23   & \textbf{85.77}   & 86.79   & \textbf{88.16}   & 89.04 & 74.33 & \textbf{89.48} & 79.69        & \textbf{80.15}        & 74.48         & \textbf{80.15}        & 84.16        & \textbf{91.43}        \\
\multicolumn{1}{c|}{Mesh}                    & Image                   & 83.58   & \textbf{85.21}   & 85.96   & \textbf{87.76}   & 88.11 & 76.18 & \textbf{88.84} & 79.14        & \textbf{82.14}        & 87.50          & \textbf{92.83}        & 94.74        & \textbf{94.81}        \\
\multicolumn{1}{c|}{Point}                   & Image                   & 82.29   & \textbf{85.23}   & 85.18   & \textbf{87.87}   & 87.11 & 73.74 & \textbf{89.03} & 79.40        & \textbf{81.82}        & 74.54         & \textbf{83.72}        & 86.22        & \textbf{93.09}        \\
\multicolumn{1}{c|}{Mesh}                    & Mesh                    & 88.51   & \textbf{89.30}   & ——      & ——               & ——    & 74.84 & ——             & 85.59        & \textbf{87.05}        & 83.36         & \textbf{90.75}        & 90.54        & \textbf{94.72}        \\
\multicolumn{1}{c|}{Mesh}                    & Point                   & 87.37   & \textbf{88.14}   & ——      & ——               & ——    & 70.21 & ——             & 85.74        & \textbf{86.59}        & 88.53         & \textbf{89.05}        & 95.22        & \textbf{95.77}        \\
\multicolumn{1}{c|}{Point}                   & Point                   & 87.04   & \textbf{88.51}   & ——      & ——               & ——    & 70.80 & ——             & 86.32        & \textbf{86.63}        & 72.58         & \textbf{81.47}        & 87.77        & \textbf{92.23}        \\
\multicolumn{1}{c|}{Point}                   & Mesh                    & 87.58   & \textbf{89.04}   & ——      & ——               & ——    & 71.59 & ——             & 85.91        & \textbf{86.85}        & 85.30          & \textbf{89.16}        & 92.43        & \textbf{92.54}        \\ \hline
\multicolumn{2}{c|}{Mean}                                              & 85.47   & \textbf{86.77}   & 86.97   & \textbf{88.39}   & 88.29 & 74.59 & \textbf{89.20} & 81.70        & \textbf{83.14}        & 79.23         & \textbf{85.20}        & 89.31        & \textbf{92.98}        \\ \hline
\end{tabular}
\label{tab:tab1}
\end{table*}

\begin{table*}[]
    \centering
        \caption{The ablation studies for loss functions on the ModelNet40 dataset. The number of views for images is fixed to 1. The same number of epochs is used for all the experiments, and batchsize is fixed to 128.}
\begin{tabular}{c|c|c|c|c|c|c|c}
\hline
Loss        & $\mathcal{L}_{CE}$    & $\mathcal{L}_{IV}$  & $\mathcal{L}_{NS'}$   & $\mathcal{L}_{IV}+\mathcal{L}_{IC}$ & $\mathcal{L}_{CE}+\mathcal{L}_{IV}$ & $\mathcal{L}_{CE}+\mathcal{L}_{IV}+\mathcal{L}_{IC}  $    & $\mathcal{L}_{CE}+\mathcal{L}_{IC}$ \\ \hline
Image2Image & 77.65 & 80.55 & 79.73 & 82.95 & 80.97 & \textbf{83.55} & 81.61 \\
Image2Mesh  & 82.26 & 84.84 & 84.54 & 85.82 & 84.90  & \textbf{86.20} & 83.05 \\
Image2Point & 76.91 & 84.09 & 83.86 & 84.86 & 83.83 & \textbf{85.77} & 81.17 \\
Mesh2Mesh   & 86.78 & 88.56 & 83.86 & 88.66 & 89.28 & \textbf{89.30} & 84.60 \\
Mesh2Image  & 82.13 & 83.19 & 82.67 & 84.93 & 83.97 & \textbf{85.21} & 82.29 \\
Mesh2Point  & 81.35 & 87.30 & 87.23 & 86.75 & 87.77 & \textbf{88.14} & 81.57 \\
Point2Point & 71.97 & 87.70 & 87.24 & 87.52 & 87.57 & \textbf{88.51} & 79.95 \\
Point2Image & 75.56 & 82.68 & 81.93 & 84.64 & 83.34 & \textbf{85.23} & 80.44 \\
Point2Mesh  & 80.83 & 88.03 & 87.88 & 88.39 & 88.32 & \textbf{89.04} & 81.96 \\ \hline
Mean        & 79.49 & 85.22 & 84.84 & 86.06 & 85.55 & \textbf{86.77} & 81.85 \\ \hline
\end{tabular}
    \label{tab:tab2}
\end{table*}

\section{Experiments}
\textbf{Datasets.} To validate our proposed method, we perform experiments on three datasets: 
% ModelNet40~\cite{DBLP:conf/cvpr/WuSKYZTX15}, MI3DOR~\cite{DBLP:conf/mm/ZhouLN19}, Pix3D~\cite{DBLP:conf/cvpr/Sun0ZZZXTF18}, and Pix3D with four categories (subset of the Pix3D dataset made by~\cite{DBLP:conf/iccv/Lin0WLJZ021}).
% The ModelNet40 dataset is a 3D object benchmark and contains 12,311 CAD models belonging to 40 different categories, with 9,843 used for training and 2,468 for testing. Three modalities are provided in this dataset, including image, point cloud,
% and mesh. MI3DOR is a large-scale dataset for 2D-to-3D tasks with 21,000 images and 7,690 models from 21 categories, with 3,842 used for training and 3,848 for testing. Pix3D is a large-scale dataset of real images and ground-truth models with precise 2D-3D alignment and contains 395 models and 16,913 images from 9 categories, with 313 used for training and 82 for testing. Pix3D with four categories(Pix3D-4) only chooses categories that contain more than 300 non-occluded and non-truncated samples. The Pix3D-4 contains 322 models from 4 categories (bed, chair, sofa, table), with 257 for training and 65 for testing. Both MI3DOR and Pix3D datasets only have 3D models (complex mesh in obj format). 所以，我们对以上两个数据集中的数据进行采样，分别得到了包括1024个面片的mesh数据和包括2048个点的点云数据 。
ModelNet40~\cite{DBLP:conf/cvpr/WuSKYZTX15}, MI3DOR~\cite{DBLP:conf/mm/ZhouLN19}, Pix3D~\cite{DBLP:conf/cvpr/Sun0ZZZXTF18}, and Pix3D with four categories (Pix3D-4, a subset of the Pix3D dataset created by~\cite{DBLP:conf/iccv/Lin0WLJZ021}).
The ModelNet40 dataset is a 3D object benchmark and contains 12,311 CAD models belonging to 40 different categories, with 9,843 used for training and 2,468 for testing. This dataset provides three modalities: image, point cloud, and mesh. MI3DOR is a large-scale dataset for 2D-to-3D tasks with 21,000 images and 7,690 models from 21 categories, with 3,842 used for training and 3,848 for testing. Pix3D is a large-scale dataset of real images and ground-truth models with precise 2D-3D alignment and contains 395 models and 16,913 images from 9 categories, with 313 used for training and 82 for testing. Pix3D-4 only chooses categories that contain more than 300 non-occluded and non-truncated samples. The Pix3D-4 contains 322 models from 4 categories (bed, chair, sofa, table), with 257 for training and 65 for testing. Both MI3DOR and Pix3D datasets only have 3D models (complex mesh in obj format).
Therefore, we sampled the 3D models from the MI3DOR and Pix3D datasets, resulting in mesh data with 1024 faces and point cloud data with 2048 points for each dataset.
\subsection{Implementation Details}
We propose an end-to-end framework for cross-modal retrieval tasks based on proposed Instance-Variant loss, Intra-Class loss, and cross-entropy loss. The overview of the 3D cross-modal retrieval task framework is illustrated in Fig. \ref{fig:fig2}. 
For 2D image feature extraction, we utilize ResNet18~\cite{DBLP:conf/cvpr/HeZRS16} as the backbone network with four convolution blocks, all with 3 × 3 kernels, where the number of kernels is 64, 128, 256, and 512, respectively. 
DGCNN is employed as the backbone network to capture point cloud features. DGCNN~\cite{DBLP:journals/tog/WangSLSBS19} contains four EdgeConv blocks with kernels set to 64, 64, 64, and 128. 
% After the four EdgeConv blocks, a fully connected layer with 512 neurons is used to extract point-specific features for each point, and then a max-pooling layer is applied to extract global features for each object.
MeshNet~\cite{DBLP:conf/aaai/FengFYZG19} is used to extract mesh features. 
The shared multimodal feature encoder consists of two fully connected layers with a size of 512,256,C (C is the number of classes). 
The three proposed loss functions are used to train the network to learn discriminative and modal-invariant features jointly:
$\mathcal{L}=\mathcal{L}_{IV}+\mathcal{L}_{IC}+\mathcal{L}_{CE}.$
% \begin{equation}
%     \mathcal{L}=\mathcal{L}_{IV}+\mathcal{L}_{IC}+\mathcal{L}_{CE}.
% \end{equation}

\textbf{Training details.}
% We implemented our network on PyTorch~\cite{DBLP:conf/nips/PaszkeGMLBCKLGA19}.
% On all three datasets, our model is trained with an SGD optimizer with a learning rate of 0.01. The learning rate is reduced by 90$\%$ every 20,000, 20,000, and 4,000 iterations for ModelNet40, MI3DOR, and Pix3D, respectively.
% We choose temperature $\omega=\frac{1}{30}$ and $\lambda_0=0.35$ for all three datasets, and $\tau=0.1$, $\tau=0.1$, $\tau=8$ for ModelNet40, MI3DOR, and Pix3D respectively. 
% Since our work is the first to employ real-world datasets like MI3DOR and Pix3D for 3D cross-modal retrieval tasks, we lack baseline methods for comparison. As a result, we have chosen cross-modal center loss~\cite{DBLP:conf/cvpr/JingVTT21} as our baseline and conducted relevant experiments on the aforementioned datasets. The training detail of cross-modal center loss on MI3DOR and Pix3D datasets is as same as its detail on ModeNet40.
We implemented our network using PyTorch~\cite{DBLP:conf/nips/PaszkeGMLBCKLGA19}. For all three datasets, our network is trained with an SGD optimizer and a learning rate of 0.01. The learning rate is reduced by 90$\%$ every 20,000 iterations for ModelNet40 and MI3DOR, and every 4,000 iterations for Pix3D. We set the temperature parameter $\omega=\frac{1}{30}$ and $\lambda_0=0.35$ for all three datasets, and $\tau=0.1$, $\tau=0.1$, and $\tau=8$ for ModelNet40, MI3DOR, and Pix3D, respectively.
Since our work is the first to employ real-world datasets like MI3DOR and Pix3D for 3D cross-modal retrieval tasks, we lack baseline methods for comparison. As a result, we have chosen Cross-Modal Center Loss (Short as CMCL)~\cite{DBLP:conf/cvpr/JingVTT21} as our baseline and conducted relevant experiments on the aforementioned datasets. CMCL training details on MI3DOR and Pix3D datasets are the same as those on ModelNet40.

\textbf{Evaluation Metrics.}
The evaluation results for all experiments are presented with the Mean Average Precision (mAP) score, a classical performance evaluation criterion for cross-modal retrieval tasks~\cite{DBLP:conf/cvpr/JingVTT21,DBLP:conf/bmvc/ChenJLT021,DBLP:conf/mm/ZengSM21}. The mAP for the retrieval task measures whether the retrieved data belong to the same class as the query (relevant) or not (irrelevant). Given a query and a set of R corresponding retrieved data (R top-ranked data), the Average Precision is defined as: 
\begin{equation}
A P=\frac{1}{T} \sum_{r=1}^R P_r \times \delta(r),
\end{equation}
where $T$ is the number of relevant items in the retrieved set, $P_r$ represents the precision of the top $r$ retrieved items, and $\delta(r)$ is an indicator function whose value is one if the $r$-th retrieved item is relevant (here relevant means belonging to the category of the query). The MAP can be calculated by averaging the AP values.
\subsection{3D Cross-modal Retrieval Task}
To evaluate the effectiveness of the proposed loss function, we conduct experiments on the ModelNet40, MI3DOR, and Pix3D datasets with three different modalities, including images, point clouds, and meshes. To thoroughly examine the quality of the learned features, we perform two retrieval tasks: uni-modal retrieval and cross-domain retrieval. The performance of our method for 3D uni-modal and cross-modal retrieval tasks is shown in Table \ref{tab:tab1}.
Since only the Cross-Modal Center Loss (CMCL) is proposed for 3D cross-modal retrieval, and they only experiment on the ModelNet dataset, we reproduce the CMCL method on MI3DOR and Pix3D datasets to evaluate the retrieval performance from real-image. Our proposed jointly trained method significantly outperforms the state-of-the-art method on all retrieval tasks and datasets.
In particular, when the input retrieval data consists of real images, our retrieval method demonstrates a more significant performance compared to the CMCL method. This indicates that our approach can effectively handle real datasets with challenging negative examples. This can be attributed to our Instance-Variant loss assigning different penalty strengths to different instances, improving the space separator. Our method obtained significantly better performance on all retrieval pairs across all datasets, showcasing the strong generalization ability of our proposed method.

\subsection{Impact of Loss Function}

The three components of our proposed loss function are as follows: Cross-entropy loss for each modality in the label space, denoted as $\mathcal{L}_{CE}$; Instance-Variant loss in the shared embedding space, denoted as $\mathcal{L}_{IV}$; and Intra-Class loss, denoted as $\mathcal{L}_{IC}$. We also mention the softmax loss without the instance-variant weight, denoted as $\mathcal{L}_{NS'}$. To further investigate the impact of each component, we evaluate different combinations for the loss functions, including optimization with $\mathcal{L}_{CE}$, $\mathcal{L}_{IV}$, and $\mathcal{L}_{NS'}$ respectively; jointly optimization with $\mathcal{L}_{CE} \& \mathcal{L}_{IV}$, $\mathcal{L}_{CE} \& \mathcal{L}_{IC}$, $\mathcal{L}_{IV} \& \mathcal{L}_{IC}$; jointly optimization with $\mathcal{L}_{CE}$, $\mathcal{L}_{IC}$, and $\mathcal{L}_{IV}$. These five networks are trained with the same setting and hyper-parameters, where the performance is shown in Table \ref{tab:tab2}. As illustrated in Table \ref{tab:tab2}. A) The combination of $\mathcal{L}_{CE}$, $\mathcal{L}_{IC}$, and $\mathcal{L}_{IV}$ achieves the best performance for all cross-modal and uni-modal retrieval tasks. B) As the baseline, cross-entropy loss alone achieves relatively high mAP due to the sharing head of the three modalities forcing the network to learn similar representations in the common space for different modalities of the same class. C) The Instance-Variant loss can be used independently, achieving fairly good retrieval results. When combined with the Intra-Class loss, some tasks' retrieval performance surpasses the CMCL method's. D)  $\mathcal{L}_{IC}$ can improve both performances of the uni-modal or cross-modal retrieval. E) $\mathcal{L}_{IV}$ is better than the $\mathcal{L}_{NS'}$.
% softmax loss without the instance-variant weight.

\begin{table}[]
    \centering
        \caption{The ablation studies for the batch size on the ModelNet40 dataset. The number of views for images is fixed to 1.}
    \label{tab:tab3}
\begin{tabular}{c|c|c|c|c}
\hline
Batchsize   & 32    & 64             & 96    & 128            \\ \hline
Image2Image & 81.78 & 83.34          & 83.45 & \textbf{83.55} \\
Image2Mesh  & 85.16 & \textbf{86.23} & 86.18 & 86.20          \\
Image2Point & 83.81 & 85.29          & 85.51 & \textbf{85.77} \\
Mesh2Mesh   & 89.57 & \textbf{89.59} & 89.47 & 89.30          \\
Mesh2Image  & 84.91 & \textbf{85.49} & 85.33 & 85.21          \\
Mesh2Point  & 87.39 & 87.95          & 87.89 & \textbf{88.14} \\
Point2Point & 86.77 & 87.49          & 88.02 & \textbf{88.51} \\
Point2Image & 83.93 & 84.84          & 85.11 & \textbf{85.23} \\
Point2Mesh  & 88.07 & 88.61          & 88.83 & \textbf{89.04} \\ \hline
Mean        & 85.71 & 86.54          & 86.64 & \textbf{86.77} \\ \hline
\end{tabular}
\end{table}

\begin{table}[]
  \centering
  \caption{Performance of 3D uni-modal and cross-modal unseen retrieval task between MI3DOR and Pix3D.}
    \begin{tabular}{c|cc|cc}
    \hline
    Unseen & \multicolumn{2}{c|}{MI3DOR: img->view} & \multicolumn{2}{c}{MI3DOR->Pix3D-9} \\
    \hline
    Method & CMCL  & Ours  & CMCL  & Ours \\
    \hline
    Image2Image & 24.77 & \textbf{30.24} & 54.15 & \textbf{54.29} \\
    Image2Mesh & 25.82 & \textbf{31.69} & \textbf{53.03} & 48.27 \\
    Image2Point & 25.92 & \textbf{31.29} & 51.14 & \textbf{53.08} \\
    Mesh2Mesh & 85.59 & \textbf{87.05} & 76.72 & \textbf{76.75} \\
    Mesh2Image & 22.45 & \textbf{25.28} & \textbf{46.32} & 42.22 \\
    Mesh2Point & 85.71 & \textbf{86.59} & 48.24 & \textbf{48.91} \\
    Point2Point & 86.19 & \textbf{86.63} & 77.75 & \textbf{80.32} \\
    Point2Image & 21.81 & \textbf{24.16} & 46.07 & \textbf{48.55} \\
    Point2Mesh & 85.80  & \textbf{86.85} & 64.17 & \textbf{64.57} \\
    \hline
    Mean  & 51.56 & \textbf{54.42} & \textbf{57.51} & 57.44 \\
    \hline
    \end{tabular}%
  \label{tab:tab4}%
\end{table}%

\subsection{Impact of Batch Size}
\begin{figure*}
    \centering
    \includegraphics[width= 0.95 \textwidth]{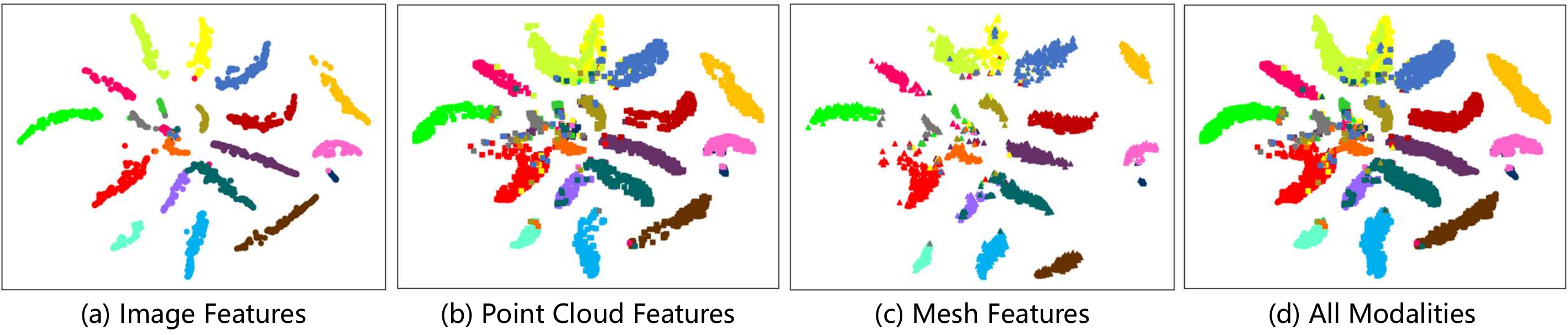}
    \caption{The visualization for the testing data in the MI3DOR dataset by using t-SNE method~\cite{van2008visualizing}. 
    % Each point in the figure represents one instance. 
    Instances from the same category are rendered with the same color.
}
    \label{fig:tsne}
\end{figure*}
\begin{figure*}
    \centering
    \includegraphics[width=\textwidth]{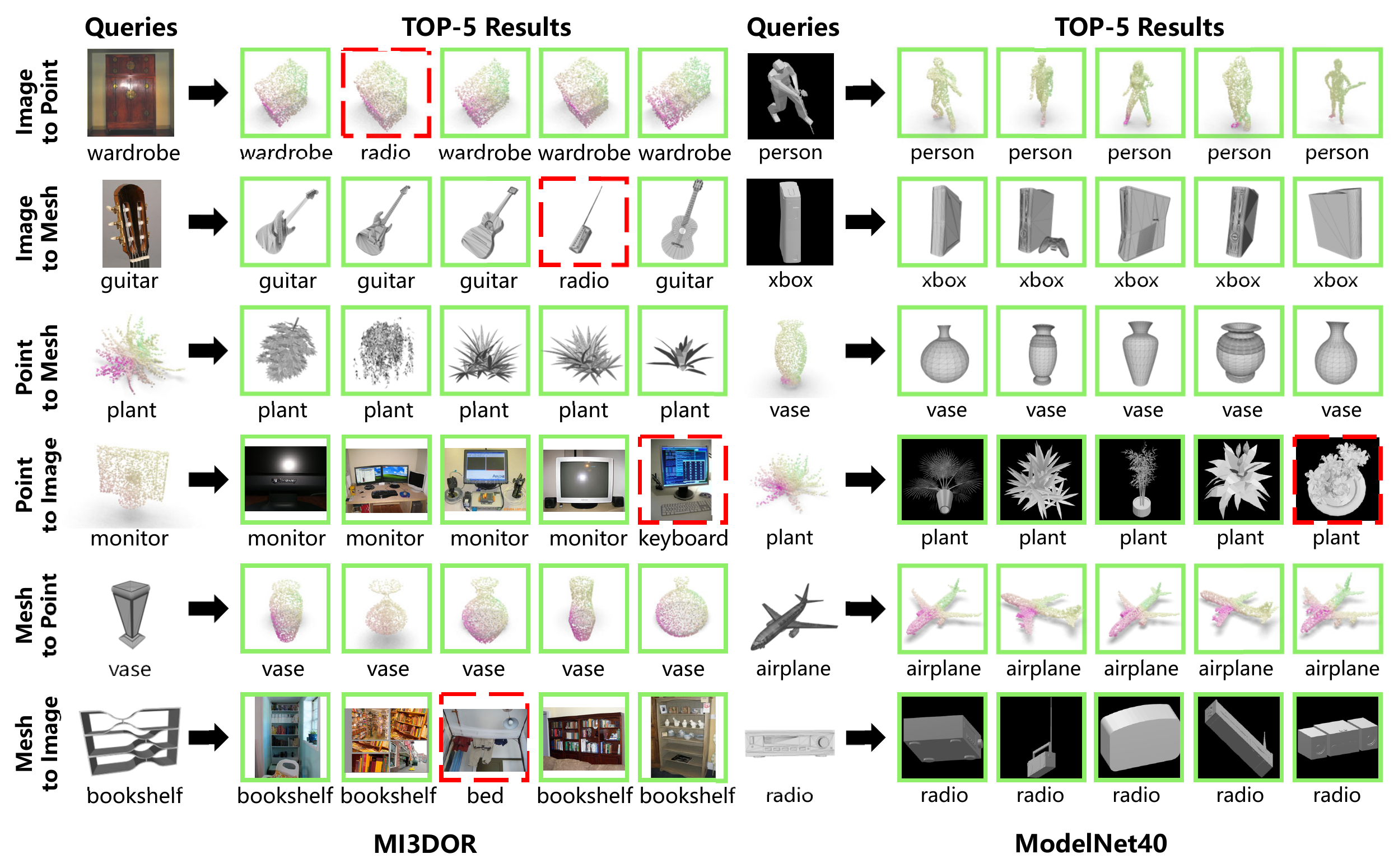}
    \caption{Top-5 retrieval results on ModelNet40 and MI3DOR datasets. The green bounding boxes indicate that the images belong to the same category as the query, whereas the red dash bounding boxes indicate wrong matches.  }
    \label{fig:v1}
\end{figure*}

Few researchers have discussed the difference between the weight vector and class center~\cite{DBLP:conf/mm/WangXCY17,DBLP:journals/spl/WangCLL18}. 
% They compared the distribution differences of the weight vector and class center before and after full optimization. They concluded that when the network optimization reaches optimum, the distribution of class centers and weight vectors will overlap~\cite{DBLP:conf/mm/WangXCY17}. We noticed that the batch size affects the retrieval performance of center loss. Therefore, we decided to explore further whether the batch size influences the weight vector calculation and whether the Intra-Class loss proposed in this paper is affected by the batch size. 
The researchers compared weight vector and class center distribution differences before and after full optimization, concluding that their distributions overlap at optimum network optimization~\cite{DBLP:conf/mm/WangXCY17}. 
It has been proved that batch size affects center loss retrieval performance~\cite{DBLP:conf/cvpr/JingVTT21,DBLP:conf/eccv/WenZL016}
Consequently, we further investigated the impact of batch size on weight vector calculation and Intra-Class loss.
To analyze the impact of batch sizes on the performance, we conduct experiments on the ModelNet40 dataset with different batch sizes (32,64,96,128). All networks are trained with the same number of epochs and hyper-parameters. As shown in Table \ref{tab:tab3}, changing the batch size does not significantly impact the retrieval performance, indicating that the weight vector and Intra-Class loss are not affected by the variations in the amount of data within a batch. This also implies that the loss function proposed in this paper can achieve comparable retrieval results with fewer resources. Meanwhile, we noticed that the performance of some retrieval tasks was improved when using smaller batch sizes. 
We speculate that the primary factors are hyperparameter selection and imbalanced modality optimization issues leading to imperfect feature extraction~\cite{DBLP:conf/cvpr/PengWD0H22}.
% We speculate that the main reasons for this are the choice of hyperparameters and the imperfect feature extraction caused by imbalanced modality optimization issues~\cite{DBLP:conf/cvpr/PengWD0H22}.
\subsection{3D Cross-modal Unseen Retrieval Task}

Considering the real-world application of cross-modal retrieval, we may encounter new data domains or distributions that have not appeared in the training dataset. Therefore, we use real-image of the MI3DOR dataset training the retrieval network, using views of MI3DOR and Pix3D datasets to test. 
The main reason for choosing the MI3DOR dataset as the benchmark is that it contains both real images corresponding to the models and multi-view images of the models; at the same time, the data distribution in this dataset is more balanced compared to Pix3D, so we do not need to worry about the unbiased nature of the networks used for testing.
As illustrated in Table \ref{tab:tab4}, our network has better generalization ability than the CMCL method. However, it fails to address the challenge of unseen domain retrieval, as the network's performance experiences a significant decline.
We will discuss more experimental results about unseen retrieval in the appendix.
% Initially, we suspected that the decline in retrieval performance was due to different category divisions within the dataset. However, when using the same modality data of different types within the same dataset for retrieval (image and view from MI3DOR), the performance still experiences a noticeable decline, indicating that the performance drop is not entirely due to the unseen classes.
% We will discuss more experimental results concerning unseen domains and unseen classes in the appendix.

\subsection{Qualitative Visualization}
\textbf{T-SNE Feature Embedding Visualization.} 
Fig. \ref{fig:tsne} demonstrates distinct clusters for each modality, highlighting the proposed approach's effectiveness in discriminating class samples. Furthermore, the combined features across modalities confirm the learned common space can capture modality-invariant representations.
% Figure \ref{fig:tsne} shows that the features are distributed as distinct clusters, demonstrating that the proposed loss can effectively discriminate samples from different classes for each modality. 
% Additionally, after combining the features from the three modalities, it is evident that the features learned by the proposed framework in the universal space are indeed modality-invariant. 

% \begin{figure}
%     \centering
%     \includegraphics[width=\columnwidth]{samples/figure/Figure5.png}
%     \caption{Top-10 retrieval results on MI3DOR dataset.  }
%     \label{fig:v2}
% \end{figure}

\textbf{3D Cross-Modal Retrieval Visualization.} Fig. \ref{fig:v1} displays the cross-modal retrieval samples for six queries from the ModelNet40 and MI3DOR datasets.  
Cosine distance measures data similarity across different modalities using normalized features for each query. 
The figure shows that instances with similar appearances are closer in feature space despite different modalities, proving the network learned modality-invariant features.
% For each query, the cosine distance over the normalized features is used to measure data similarity from different modalities. 
% The figure demonstrates that objects with similar appearances are closer in the feature space, even though they are from different modalities, proving that the network learned modality-invariant features. 
More experiment results will be illustrated in the appendix.
% We will illustrate more experiment results in our appendix.

% Due to the data distribution in the Pix3D dataset, where some classes have only 3-4 test samples, we cannot calculate the top-10 results for the Pix3D dataset. Therefore, we only visualize the retrieval results for the ModelNet40 and MI3DOR datasets.

\section{Conclusion}
This paper introduces the Instance-Variant loss and Intra-Class loss for 3D cross-modal retrieval. 
The Instance-Variant loss effectively assigns different penalty strengths to different instances, enhancing the network's effectiveness by focusing on more challenging instances and promoting better feature space separability.
The Intra-Class loss, based on the Gaussian RBF kernel, aims to minimize the intra-class distance among all instances in the shared embedding space. This approach minimizes the intra-class distance and evaluates shared embedding space learned by the network. 
The proposed Instance-Variant loss learns a shared weight vector for data from different modalities, effectively mitigating cross-modal discrepancy and enhancing cross-modal retrieval performance.
Extensive experiments have been conducted on 3D cross-modal retrieval tasks. The proposed framework significantly outperforms the state-of-the-art methods on the ModelNet40, MI3DOR, and Pix3D datasets. 
In future work, we will explore potential technologies further to improve the robustness and effectiveness of the proposed loss and enhance the unseen retrieval performance.

\bibliographystyle{ACM-Reference-Format}
\bibliography{sample-base}


\end{document}